\documentclass[9pt]{article}

\usepackage{hslTR}

\usepackage{graphics}
\usepackage{graphicx}
\usepackage{verbatim}     
\usepackage{amsxtra}
\usepackage{subfigure}
\usepackage{hyperref} 
\usepackage{amssymb}
\usepackage{amsmath}
\usepackage{newtxsf}
\usepackage{ifthen}
\usepackage{color}
\usepackage{float}
\usepackage{mathtools}
 \usepackage{enumitem}
\usepackage{algorithm}
\usepackage{algorithmic}
\usepackage{listings}

\usepackage{rgsEnvironments}
\usepackage{rgsMacros}

\definecolor{mygreen}{RGB}{0,128,0}
\definecolor{purple}{RGB}{128,0,128}
\usepackage{xcolor}

\definecolor{codegreen}{rgb}{0,0.6,0}
\definecolor{codegray}{rgb}{0.5,0.5,0.5}
\definecolor{codepurple}{rgb}{0.58,0,0.82}
\definecolor{backcolour}{rgb}{0.95,0.95,0.92}
\definecolor{blue}{rgb}{0, 0, 1}
\newbool{shorten}
\setbool{shorten}{false}

\lstdefinestyle{mystyle}{
    commentstyle=\color{codegreen},
    keywordstyle=\color{purple},
    numberstyle=\tiny\color{codegray},
    stringstyle=\color{codepurple},
    basicstyle=\sffamily\footnotesize,
    breakatwhitespace=false,         
    breaklines=true,                 
    captionpos=b,                    
    keepspaces=true,                 
    numbers=left,                    
    numbersep=5pt,                  
    showspaces=false,                
    showstringspaces=false,
    showtabs=false,                  
    tabsize=2
}

\usepackage{import}

\newcommand{\nw}[1]{{\color{black}#1}}
\newcommand{\revised}[1]{\color{black}#1}
\newcommand{\revision}[1]{\color{black}#1}
\newcommand{\revisions}[1]{\color{black}#1}
\newcommand\BibTeX{{\rmfamily B\kern-.05em \textsc{i\kern-.025em b}\kern-.08em
T\kern-.1667em\lower.7ex\hbox{E}\kern-.125emX}}

\DeclareMathOperator{\interior}{int}

\newcommand{\nnumbers}[1][]{\mathbb{N}}

\newcommand{\Function}[2]{\STATE \textbf{function} #1\textnormal{(#2)}}

\newcounter{theorem}
\newtheorem{problem}[theorem]{Problem}

\newlist{steps}{enumerate}{1}
\setlist[steps, 1]{leftmargin=45pt, label = \textbf{Step \arabic*}:}

\usepackage[square, numbers]{natbib}
\title{\LARGE \bf
cHyRRT and cHySST: Motion Planning Tools for Hybrid Dynamical Systems in OMPL}

\author{Beverly Xu, Nan Wang, and Ricardo G. Sanfelice$^{1}$
\thanks{*Research by B. Xu, N. Wang and R. G. Sanfelice is partially supported by NSF Grants no. CNS-2039054 and CNS-2111688, by AFOSR Grants nos. FA9550-19-1-0169, FA9550-20-1-0238, FA9550-23-1-0145, and FA9550-23-1-0313, by AFRL Grant nos. FA8651-22-1-0017 and FA8651-23-1-0004, by ARO Grant no. W911NF-20-1-0253, and by DoD Grant no. W911NF-23-1-0158.}
\thanks{$^{1}$Beverly Xu, Nan Wang, and Ricardo G. Sanfelice are with the Department of Computer Engineering, University of California, Santa Cruz, Santa Cruz 95064, CA 
        {\tt\small xu21beve@gmail.com, nanwang@ucsc.edu, ricardo@ucsc.edu}}%
\thanks{$^{1}$Beverly Xu was a research intern at HSL during this work.}%
}

\begin{document}

\maketitle
\thispagestyle{empty}
\pagestyle{empty}

\begin{revised}
\begin{abstract}
%
This paper presents two implementations of the recently developed motion planning algorithms HyRRT \cite{wang2022rapidly} and HySST \cite{wang2023hysst}. Specifically, cHyRRT, an implementation of the HyRRT algorithm, generates solutions to motion planning problems for hybrid systems with a probabilistic completeness guarantee, while cHySST, an implementation of the asymptotically near-optimal HySST algorithm, finds near-optimal trajectories based on a user-defined cost function.
The implementations align with the theoretical foundations of hybrid system theory and are designed based on OMPL, ensuring compatibility with ROS while prioritizing computational efficiency. The structure, components, and usage of both tools are detailed. A modified pinball game and collision-resilient tensegrity multicopter example are provided to illustrate the tools' key capabilities.
\end{abstract}
\end{revised}

\section{Introduction}

Motion planning is becoming an increasingly important tool for researchers and practitioners as emerging robotic systems such as quadrupled robots and drones enter operation in complex environments. 
Among various implementations, the Open Motion Planning Library (OMPL) \cite{sucan2012the-open-motion-planning-library} is widely used due to its compatibility with the Robot Operating System (ROS) and its comprehensive collection of state-of-the-art planners for systems with purely continuous-time or purely discrete-time dynamics. However, motion planning for hybrid systems, in which the states can evolve continuously and, at times, exhibit jumps, along with implementations of such algorithms, are not currently available.

{We consider the hybrid equation framework in \cite{goebel2009hybrid} to model hybrid dynamical systems,}
\begin{equation}
\mathcal{H}: \left\{              
\begin{aligned}               
\dot{x} = f(x, u) \quad (x, u)\in C\\               
x^{+} = g(x, u) \quad (x, u)\in D\\                
\end{aligned}   
\right.
\label{model:generalhybridsystem}
\end{equation}
where $x\in \mathbb{R}^n$ is the state, $u\in \mathbb{R}^m$ is the input, $C\subset \mathbb{R}^{n}\times\mathbb{R}^{m}$ represents the flow set, $f: \mathbb{R}^{n}\times\mathbb{R}^{m} \to \mathbb{R}^{n}$ represents the flow map, $D\subset \mathbb{R}^{n}\times\mathbb{R}^{m}$ represents the jump set, and $g:\mathbb{R}^{n}\times\mathbb{R}^{m} \to \mathbb{R}^{n}$ represents the jump map. Here, flow map $f$ and jump map $g$ capture continuous and discrete evolution of $x$, respectively. The flow set $C$ collects the points where the state can evolve continuously. The jump set $D$ collects the points where jumps can occur. This general framework for hybrid systems captures a broad class of hybrid systems, including the class of hybrid systems considered in \cite{branicky2003samplingbased} and those that incorporate timers, impulses, constraints, and environmental contacts.

Within this framework, the feasible and the optimal motion planning problems for hybrid dynamical systems are formulated and addressed in \cite{wang2022rapidly} and \cite{wang2023hysst}, respectively. In practice, Rapidly-exploring Random Trees (RRT)-type algorithms efficiently compute trajectories for high-dimensional problems by incrementally constructing a search tree through random state-space sampling. Following the RRT algorithm scheme, HyRRT in \cite{wang2022rapidly} solves the feasible motion planning problem for hybrid systems, inheriting its probabilistically complete guarantee, meaning that the probability of failing to find a motion plan converges to zero as the number of samples approaches infinity \cite{LaValle1998RapidlyexploringRT}. 
\ifbool{shorten}{}{At each iteration, HyRRT randomly picks a state sample, selects the vertex such that the state associated with this vertex has minimal distance to the sample, and extends the search tree by flow or jump, which is also chosen randomly when both regimes are possible. 
Therefore, the planner takes in flow and jump maps $f$ and $g$ and flow and jump sets $C$ and $D$ representing the system’s dynamics, starting, final, and unsafe state sets, to return an OMPL solution status and OMPL motion plan.}

In many motion planning applications, an optimal solution is preferred over merely a feasible but suboptimal one \cite{hysst-6}. Unfortunately, solutions generated by RRT may converge to a suboptimal plan \cite{hysst-7}, while optimal variants such as PRM* and RRT* \cite{hysst-8} rely on steering functions, restricting applicability. In contrast, the stable sparse RRT (SST) algorithm \cite{hysst-9} eliminates the need for a steering function while guaranteeing asymptotic near-optimality, meaning that as the number of samples approaches infinity, the probability of finding a solution with a cost close to the minimum converges to one. Therefore, HySST is particularly useful when a user-defined cost functional is provided to evaluate the quality of solutions.
HySST differs from HyRRT in two aspects: i) HySST takes in the input pruning radius $\delta_{S} \in \mathbb{R}_{>0} $ to remove all vertices, excluding the vertex with the lowest cost, within $\delta_{S}$ of the (static) witness state, and ii) HySST takes in the input selection radius $\delta_{BN} \in \mathbb{R}_{>0}$ to select the vertex with the lowest cost within $\delta_{BN}$ of the randomly sampled vertex to start propagation from. 
If there are no vertices within the ball defined by radius $\delta_{BN}$, then the nearest vertex is selected. 
To date, no motion planners for hybrid dynamical systems have been implemented within OMPL. \ifbool{shorten}{}{Such planners are highly valuable as OMPL provides benchmarking tools and is compatible with the widely used ROS \cite{doi:10.1126/scirobotics.abm6074}, both directly and through MoveIt 2\footnote{See https://moveit.ai/ for more information about the MoveIt library.}, a widely-used robotic manipulation software and the core manipulation platform for ROS 2.  Compatibility with ROS, which provides libraries and tools for developing, simulating, and visualizing robots, is a key advantage of both tools. However, those features are not available for hybrid dynamical systems.} To address this gap, we introduce cHyRRT and cHySST\footnote{Code for both cHyRRT and cHySST is available at \href{https://github.com/xu21beve/ompl}{https://github.com/xu21beve/ompl}}, two planners integrated into OMPL, allowing for compatibility with ROS~\cite{doi:10.1126/scirobotics.abm6074}.
%
%

\ifbool{shorten}{}{The remainder of the paper is structured as follows. Section 2 presents notation and preliminaries. Section 3 presents the motion planning problem for hybrid dynamical systems. Section 4 presents the HyRRT algorithm and details of its implementation, usage, and customizations within cHyRRT. Section 5 presents similar details for HySST algorithm. Section 6 presents example applications of both algorithms and illustrations of the generated motion plans. \nw{Section 7 presents an analysis of costs for plans generated by HySST and a discussion of limitations of both algorithms.}} 

\section{{Notation and Preliminaries}}
\subsection{Notation}
The set of real numbers is denoted as $\mathbb{R}$ and its nonnegative subset is denoted as $\mathbb{R}_{\geq 0}$. The set of nonnegative integers is denoted as $\mathbb{N}$. The notation $\interior I$ denotes the interior of the interval $I$. Given sets \( P \subseteq \mathbb{R}^n \) and \( Q \subseteq \mathbb{R}^n \), the Minkowski sum of \( P \) and \( Q \), denoted as \( P + Q \), is the set \( \{ p + q : p \in P, q \in Q \} \). The notation \(\mathrm{rge} \, f\) denotes the range of the function \(f\). The notation \( |\cdot|\) denotes a norm. \ifbool{shorten}{Given a vector $h\in \mathbb{R}^2$, $h_x$ is its first component, and $h_y$ is its second component.}{}

\subsection{Preliminaries}
Given a flow set \( C \), the set \( U_C := \{ u \in \mathbb{R}^m : \exists x \in \mathbb{R}^n \texttt{ such that } (x, u) \in C \} \) includes all possible input values that can be applied during flows. Similarly, given a jump set \( D \), the set \( U_D := \{ u \in \mathbb{R}^m : \exists x \in \mathbb{R}^n \texttt{ such that } (x, u) \in D \} \) includes all possible input values that can be applied at jumps. These sets satisfy \( C \subseteq \mathbb{R}^n \times U_C \) and \( D \subseteq \mathbb{R}^n \times U_D \). Given a set \( K \subseteq \mathbb{R}^n \times U_\star \), where \( \star \) is either \( C \) or \( D \), we define \( \pi_\star(K) := \{ x : \exists u \in U_\star \texttt{ such that } (x, u) \in K \} \) as the projection of \( K \) onto \( \mathbb{R}^n \), and define \( C' := \pi_C (C) \) and \( D' := \pi_D(D) \).

In addition to ordinary time \( t \in \mathbb{R}_{\geq 0} \), we employ \( j \in \mathbb{N} \) to denote the number of jumps of the evolution of \( x \) and \( u \) for $\mathcal{H}$ in (1), leading to hybrid time \( (t, j) \) for the parameterization of its solutions and inputs. The domain of a solution to $\mathcal{H}$ is given by a hybrid time domain. A hybrid time domain is defined as a subset \( E \) of \( \mathbb{R}_{\geq 0} \times \mathbb{N} \) that, for each \( (T, J) \in E \), \( E \cap ([0, T] \times \{ 0, 1, \ldots, J \}) \) can be written as \( \bigcup_{j=0}^{J}([t_j, t_{j+1}], j) \) for some finite sequence of times \( 0 = t_0 \leq t_1 \leq t_2 \leq \ldots \leq t_{J+1} = T \). A hybrid arc \( \phi : \mathrm{dom} \, \phi \rightarrow \mathbb{R}^n \) is a function on a hybrid time domain that, for each \( j \in \mathbb{N} \), \( t \mapsto \phi(t, j) \) is locally absolutely continuous on each interval \( I^j := \{ t : (t, j) \in \mathrm{dom} \, \phi \} \) with nonempty interior. 

The definition of a solution pair to a hybrid system is given as follows.
\begin{definition}[Solution pair to a hybrid system\label{definition:solution-pair}]~Given a pair of functions \(\phi : \mathrm{dom} \, \phi \rightarrow \mathbb{R}^n\) and \(u : \mathrm{dom} \, u \rightarrow \mathbb{R}^m\), \((\phi, u)\) is a solution pair to (1) if \(\mathrm{dom}(\phi, u) := \mathrm{dom} \, \phi = \mathrm{dom} \, u\) is a hybrid time domain, \((\phi(0, 0), u(0, 0)) \in C \cup D\), and the following hold:
\begin{enumerate}
    \item[1)] For all \(j \in \mathbb{N}\) such that \(I^j\) has nonempty interior,
    \begin{enumerate}
        \item[a)] the function \(t \mapsto \phi(t, j)\) is locally absolutely continuous over \( I^j \),
        \item[b)] \((\phi(t, j), u(t, j)) \in C\) for all \(t \in \texttt{int} \, I^j\),
        \item[c)] the function \(t \mapsto u(t, j)\) is Lebesgue measurable and
        locally bounded,
        \item[d)] for almost all \(t \in I^j\), \(\dot{\phi}(t, j) = f(\phi(t, j), u(t, j))\).
    \end{enumerate}
    \item[2)] For all \((t, j) \in \mathrm{dom}(\phi, u)\) such that \((t, j + 1) \in \mathrm{dom}(\phi, u)\), \[(\phi(t, j), u(t, j)) \in D \quad \phi(t, j + 1) = g(\phi(t, j), u(t, j))\]
\end{enumerate}
\end{definition}

Our motion planning algorithms require concatenating solution pairs. The concatenation operation of solution pairs is defined next.
\begin{definition}[Concatenation operation] \label{definition:concatenation-operation}
 Given two functions $\phi_1 : \texttt{dom } \phi_1 \to \mathbb{R}^n$ and $\phi_2 : \texttt{dom } \phi_2 \to \mathbb{R}^n$, where $\texttt{dom } \phi_1$ and $\texttt{dom } \phi_2$ are hybrid time domains, $\phi_2$ can be concatenated to $\phi_1$ if $\phi_1$ is compact and $\phi : \texttt{dom } \phi \to \mathbb{R}^n$ is the concatenation of $\phi_2$ to $\phi_1$, denoted $\phi = \phi_1 \vert \phi_2$, namely:
\begin{enumerate}
    \item[1)] $\texttt{dom } \phi = \texttt{dom } \phi_1 \, \cup \, (\texttt{dom } \phi_2 + \{(T, J)\})$, where $(T, J) = \max \texttt{dom } \phi_1$ and the plus sign denotes Minkowski addition;
    \item[2)] $\phi(t, j) = \phi_1(t, j)$ for all $(t, j) \in \texttt{dom } \phi_1 \setminus \{(T, J)\}$ and $\phi(t, j) = \phi_2(t - T, j - J)$ for all $(t, j) \in \texttt{dom } \phi_2 + \{(T, J)\}$.
\end{enumerate}
\end{definition}

\subsection{C++ Notation}
\begin{revision}
{In this paper, the following C++ notation is used. 
\begin{revised} Namespace is a declarative region providing scope to one or more identifiers. In this paper, we also assume use of \texttt{namespace ob = ompl::base} and \texttt{namespace oc = ompl::control}. \end{revised}
A lambda function is an anonymous function object that can be passed in as an argument.
A pointer is a variable storage of the memory address of another variable. An asterisk *, when preceding a variable name at the time of declaration, is used to declare a pointer.
The sequence of symbols $\texttt{::}$ is the scope resolution operator used to traverse scopes such as namespaces and classes, to access identifiers.
A double-valued scalar is a real, floating-point number with a maximum size of $64$ bits.
}

\begin{revised}
\section{Motion Planning for Hybrid Dynamical Systems}
This section formulates the motion planning problem for hybrid dynamical systems following \cite{wang2023hysst} and presents the data structure that implements its solution.
 \subsection{Problem Formulation}
This paper solves a motion planning problem for hybrid dynamical systems, defined by flow and jump sets $C$ and $D$, flow and jump maps $f$ and $g$, along with the conditions to i) start from a given initial state, ii) end within a given goal state set, and iii) avoid reaching the unsafe set. This problem is mathematically formulated as follows. 
\ifbool{shorten}{}{
\begin{figure}[htbp]
    \centering
\def\svgwidth{0.5\columnwidth}
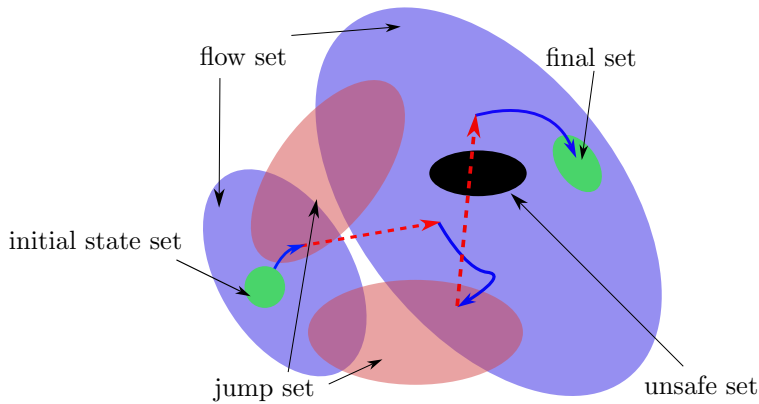

    \caption{Illustration of a sample motion plan to Problem \ref{planning-problem}, where the solid blue lines denote flow and dotted red lines denote jumps in the motion plan.} \label{fig:1}
\end{figure}
}
\end{revised}
\end{revision}
\begin{problem} \label{planning-problem}
Given a hybrid system $\mathcal{H}$ with input \( u \in \mathbb{R}^m \)
and state \( x \in \mathbb{R}^n \), the initial state set \( X_0 \subset \mathbb{R}^n \), the final
state set \( X_f \subset \mathbb{R}^n \), and the unsafe set \( X_u \subset \mathbb{R}^n \times \mathbb{R}^m \), find
a pair \((\phi, u) : \mathrm{dom}(\phi, u) \rightarrow \mathbb{R}^n \times \mathbb{R}^m\), namely a \emph{motion
plan} \ifbool{shorten}{}{in Fig. \ref{fig:1}}, such that for some \((T, J) \in \mathrm{dom}(\phi, u)\), the following
hold:
\begin{enumerate}
    \item[1)] \( \phi(0, 0) \in X_0 \), namely, the initial state of the solution
    belongs to the given initial state set \( X_0 \);
    \item[2)] \((\phi, u)\) is a solution pair to $\mathcal{H}$ as defined in Definition \ref{definition:solution-pair};
    \item[3)] \((T, J)\) is such that \( \phi(T, J) \in X_f \), namely, the solution
    belongs to the final state set at hybrid time \((T, J)\);
    \item[4)] \((\phi(t, j), u(t, j)) \notin X_u \) for each \((t, j) \in \mathrm{dom}(\phi, u)\) such
    that \( t + j \leq T+ J \), namely, the solution pair does not
    intersect with the unsafe set before its state trajectory
    reaches the final state set.
\end{enumerate}
Therefore, given sets \( X_0 \), \( X_f \), and \( X_u \), and a hybrid system
$\mathcal{H}$ with data \( (C, f, D, g) \), a motion planning problem \( P \) is
formulated as \( P = (X_0, X_f, X_u, (C, f, D, g)) \).
\end{problem}

This problem is illustrated in the following example.
\begin{revised}
\begin{example}[Modified Pinball Game] \label{example:pinball}
The state of the pinball \ifbool{shorten}{consists of the position vector $p := (p_x, p_y) \in \mathbb{R}^2$, where $p_x$ denotes the position along the $x$-axis and $p_y$ denotes the position along the $y$-axis, the velocity vector $v := (v_x, v_y) \in \mathbb{R}^2$, where $v_x$ denotes the velocity along the $x$-axis and $v_y$ denotes the velocity along the $y$-axis, and the acceleration vector $a := (a_x, a_y) \in \mathbb{R}^2$ where $a_x$ denotes the acceleration along the $x$-axis and $a_y$ denotes the acceleration along the $y$-axis. The state of the system is 
}{is composed of the position vector $p := (p_x, p_y) \in \mathbb{R}^2$, the velocity vector $v := (v_x, v_y) \in \mathbb{R}^2$, and the acceleration vector $a := (a_x, a_y) \in \mathbb{R}^2$.} The state of the system is $x := (p, v, a) \in \mathbb{R}^6$ and its input is $u := u_x \in \mathbb{R} : u_x \in [-4, 4]$. 
\begin{figure}[htbp]
        \subfigure[Simulation result solved by cHyRRT with initial position  $p_0=(4, 0)$.]{\includegraphics[width=0.315\columnwidth]{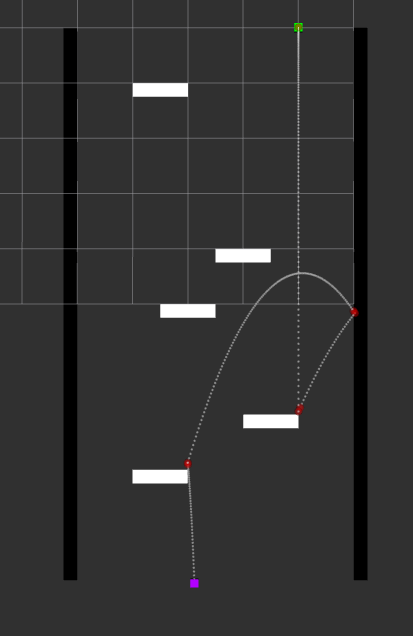}}       
        \subfigure[Simulation result solved by cHyRRT with initial position $p_0=(3.5, 0)$.]{\includegraphics[width=0.332\columnwidth]{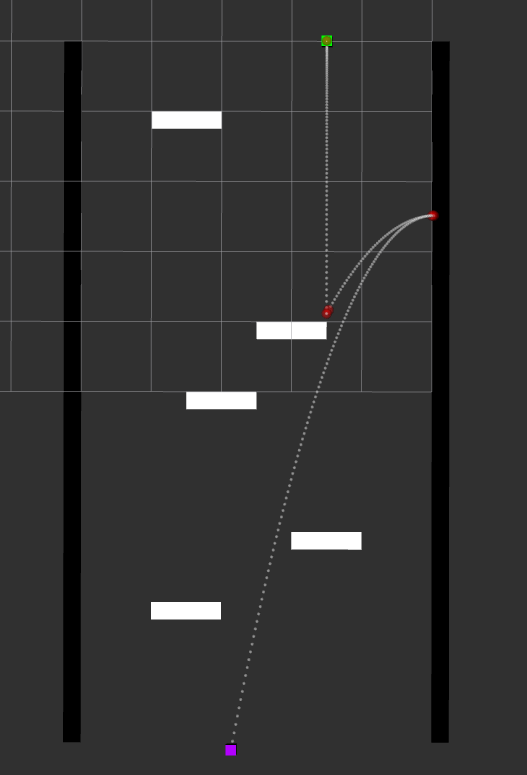}}
        \subfigure[Simulation result solved by cHySST with initial position $p_0=(3.5, 0)$.]{\includegraphics[width=0.31\columnwidth]{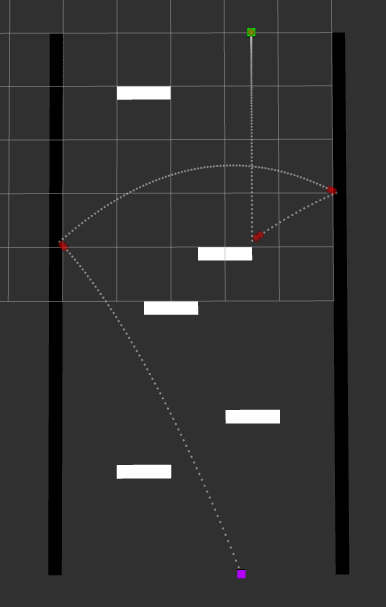}} 
        \vspace{-0.1cm}
        \caption{\revised{Simulated solution to the pinball example, graphed as $p_x$ vs $p_y$. The start and goal vertices are marked by green and purple squares, respectively, and vertices in the jump regime are marked by red circles.}}
        \label{fig:pinball}
    \end{figure}
The pinball machine environment is composed of actuated paddles, as represented by the white rectangles in Figure \ref{fig:pinball} and the two unactuated walls, represented by the black rectangles in Figure \ref{fig:pinball}. We define the upper right corner of the left wall to be $p_{origin}=(0, 0)$, the downward direction as the negative $y$, and the rightward direction as the positive $x$. The union of the paddles and the walls defines the region $M  \subset \mathbb{R}^2 $. Hence, the flow set is $C := \{((p, v, a), u) \in \mathbb{R}^6 \times \mathbb{R}^2 : p \in \mathbb{R}^2 \setminus M\}$. When the state of the pinball is in the flow set, its dynamics are governed by the following flow map:
$
\dot{x} =
\begin{bmatrix}
v \\
a \\
u
\end{bmatrix}
=: f(x, u) \quad (x, u) \in C.
$
The jump set is defined as $D := \{((p, v, a), u) \in \mathbb{R}^6 \times \mathbb{R}^2 : p \in \partial M, v \cdot n(p) \geq 0\}$, where $n(p)$ is the inward-pointing normal to the boundary $\partial M$ of the closest paddle or wall, when the pinball is at position $p$. When the pinball experiences collision with the jump set $D$,  $p^+ = p$. Denote the velocity component of $v = (v_x, v_y)$ that is normal to the wall as $v_n$ and the velocity component that is tangential to the wall as $v_t$. Then, the velocity component $v_n$ after the jump is modeled as $v_n^+ = -e v_n + u_{paddle} =: g_{v_n}(v)$ where $e \in (0, 1)$ is the coefficient of restitution. Note that the input $u_{paddle}$ for $v_n$ is only nonzero during collisions with the vertical paddle sides\ifbool{shorten}{.}{, and that the paddles and walls are characterized by distinct coefficients of restitution $e$.} During such collisions, the vectors $u_{paddle}$ and $-ev_n$ point in the same direction. 
  The velocity component $v_t$ after the jump is modeled as $v_t^+ = v_t + u =: g_{v_t}(v)$ where the input $u$ for $v_t$ is only nonzero during collisions with the top surface of a paddle.  Denote the projection of the updated vector $(v_n^+, v_t^+)$ onto the $x$-axis as $\overline{x}(v_n^+, v_t^+)$ and the projection of the updated vector $(v_n^+, v_t^+)$ onto the $y$-axis as $\overline{y}(v_n^+, v_t^+)$. Therefore,
$
v^+ =
\begin{bmatrix}
\overline{x}(g_{v_n}(v), g_{v_t}(v)) \\
\overline{y}(g_{v_n}(v), g_{v_t}(v))
\end{bmatrix}
=: g_v(v).
$
In both flow and jump sets, the dynamics of the pinball assume $a^+ = 0$. The discrete dynamics capturing the collision process is modeled as
$
x^+ =
\begin{bmatrix}
p \\
g_v(v) \\
0
\end{bmatrix}
=: g(x, u) \quad (x, u) \in D.
$
Given the initial state set as $X_0 := \{0.5, 1, 2, 3.5, 4, 4.5\}  \times \{0\}^4 \times~\{-9.81\}$, the final state set as $X_f = [1, 4] \times \{-10\} \times \mathbb{R}^4$, and the unsafe set as
$$
\begin{aligned}
    X_u = \{(x, u) \in \mathbb{R}^6 \times \mathbb{R}^2 : p_x \in [0, 1) \cup (4, 5] \cup 
    \\ p_y \in (-\infty, -10), (p_x, p_y) \in \texttt{int}\, M\},
\end{aligned}
$$
The selection radius $\epsilon_{BN}:=0.8$ and the pruning radius $\epsilon_{S}:=0.2$.  The example motion planning problem is set as \( P = (X_0, X_f, X_u, (C, f, D, g)) \). As the paddles obstruct a no-collision solution, the pinball machine example demonstrates a key strength of HyRRT over RRT; the capability to solve motion planning problems requiring collisions. Solutions to the pinball system are presented later in this paper.
\end{example}
\end{revised}

\subsection{Data structures}
Both HyRRT and HySST algorithms search for motion plans by incrementally
constructing search trees \cite{wang2022rapidly}. The search tree is modeled by
a directed tree. A directed tree $\mathcal{T}$ is a pair \( \mathcal{T} = (V, E) \),
where \( V \) is a set whose elements are called vertices and \( E \)
is a set of paired vertices whose elements are called edges.
\begin{figure}[htbp]
    \centering
\def\svgwidth{0.8\columnwidth}
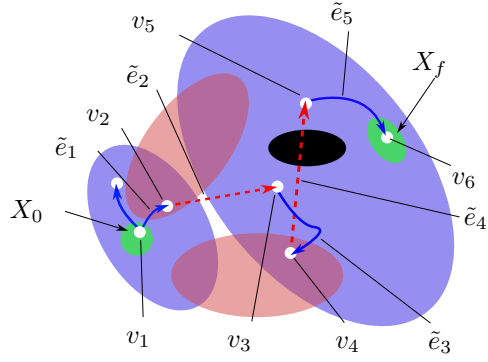

    \caption{Illustration of the search tree constructed by HyRRT/HySST. The path \( p = (v_1, v_2, \dots, v_6) \) and the solution pair \( \tilde{e}_p = \tilde{e}_1 | \tilde{e}_2 | \dots | \tilde{e}_5 \).}
\end{figure}
Each vertex in the search tree \( \mathcal{T} \) is associated with a state
value of $\mathcal{H}$. Each edge in the search tree is associated with
a solution pair to $\mathcal{H}$ that connects the state values associated
with their endpoint vertices. The state value associated with vertex \( v \in V \) is denoted as \( x_v \) and the solution pair associated with edge \( e \in E \) is denoted as \( \tilde{e} = (\phi, u)\), where \( \phi :\) dom \(\phi \rightarrow \mathbb{R}^n, u : \) dom \(u \rightarrow \mathbb{R}^m\). The solution pair that the path \( p = (v_1, v_2, \dots, v_k) \) represents is the concatenation of all those solutions associated with the edges therein, namely,
$
\tilde{p} := \tilde{e}_{(v_1, v_2)} | \tilde{e}_{(v_2, v_3)} | \dots | \tilde{e}_{(v_{k-1}, v_k)}
$
where \( \tilde{p} \) denotes the solution pair associated with the path
\( p \). 
\subsubsection{State Space} \label{state-space}
As we assume the sets $X_0, C, D, U_C,$ and $ U_D$ to have finite and positive Lebesgue measure \cite[Assumption 6.15]{wang2024motion}, we consider motion planning problems in a finite state space, which both algorithms make random selection from. We implement a state space using any derived clas of the abstract OMPL class \texttt{ob::StateSpace}\footnote{See: https://ompl.kavrakilab.org/classompl\_1\_1base\_1\_1StateSpace.html}.
To represent a state space of $n$ dimensions, constrained in each dimension by a minimum value \texttt{min\_n} and a maximum value \texttt{max\_n}, we first instantiate the state space as follows:
\begin{lstlisting}[language=C++]
ob::RealVectorStateSpace *statespace = new ob::RealVectorStateSpace(0);
\end{lstlisting}
{Then, we constrain each dimension by repeating the following with all $n$ minimum and maximum values: }
\begin{lstlisting}[language=C++]
statespace->addDimension(min_n, max_n);
\end{lstlisting}
{\revisions{The dimension of the state space corresponds to the size of the state in the hybrid model. The order in which each dimension is added corresponds to its index when accessing states sampled from the state space.}

This is illustrated in the following example. 

\begin{example}[Example \ref{example:pinball}, revisited]
We instantiate the state space of the hybrid system in Example \ref{example:pinball} as follows: 
\begin{lstlisting}[language=C++]
ob::RealVectorStateSpace *statespace = new ob::RealVectorStateSpace(0);
statespace->addDimension(min_1, max_1); // This is x_1 because it is added first
statespace->addDimension(min_2, max_2); // This is x_2 because it is added second
...
ob::HybridStateSpace *hybridSpace = new ob::HybridStateSpace(stateSpacePtr);
    ob::StateSpacePtr hybridSpacePtr(hybridSpace);

\end{lstlisting}
Then, when accessing states sampled from the state space, each dimension's order of addition corresponds to its index within the vector; for example, the first dimension to be added has a vector index of $0$; the second dimension a vector index of $1$, and so on.
\begin{lstlisting}[language=C++]
// Assuming proper instantiation of ob::State *state
double x_1 = state->as<ob::HybridStateSpace::StateType>()->as<ob::RealVectorStateSpace::StateType>(0)->values[0];
double x_2 = state->as<ob::HybridStateSpace::StateType>()->as<ob::RealVectorStateSpace::StateType>(0)->values[1];

\end{lstlisting}
\end{example}
Then, we implement the state value associated with vertex \( v \in V \), \( x_v \), as the OMPL class \texttt{ob::State}  \footnote{See OMPL class reference: https://ompl.kavrakilab.org/classompl\_1\_1base\\\_1\_1State.html}, where a state is of $n$ dimensions.}

\subsubsection{Solution Pair\label{solution-pair}}
Concatenation of solution pairs, as defined in Definition \ref{definition:concatenation-operation}, is required by our motion planning algorithms. Therefore, we augment the OMPL data structure \texttt{Motion}\footnote{See: https://ompl.kavrakilab.org/classompl\_1\_1geometric\_1\_1RRT\_1\_1\\Motion.html} to store an edge, which links the associated solution pair and inputs to other objects of \texttt{Motion} which share either endpoint of edge $e$. The augmented \texttt{Motion} class is implemented as follows:
\begin{lstlisting}[language=C++]
class Motion {
    ob::State *state;
    Motion *parent;
    std::vector<ob::State *> *solutionPair;
    oc::Control input;
    unsigned numChildren; // Only used in HySST:
    bool inactive; // Only used in HySST:
    ob::Cost accCost; // Only used in HySST:
};
\end{lstlisting}

\begin{revision}In the data structure \texttt{Motion}, the discretized solution pair \( \tilde{e} \) is implemented as a vector of states \texttt{solutionPair}, following Definition \ref{definition:solution-pair}. The input associated with each discretized state, sampled from the input sets $(U_C, U_D)$, is stored in \texttt{Motion} as the control input \texttt{input}. 
\begin{revised}
The complete implementation details for hybrid time, inputs, and edge associated with each \texttt{Motion} are introduced in the forthcoming subsections.
\end{revised}
\end{revision}

\begin{revised}
\subsubsection{Hybrid Time}
{In the \texttt{Motion} datastructure, each state $x$ is parameterized by a hybrid time \( (t, j) \), where $t$ is a double-valued scalar and $j$ is an integer-valued scalar. As this implementation depends on OMPL, which does not presently contain a state space class with the capability to capture hybrid time, we designed such a class, \texttt{ob::HybridStateSpace}, inheriting from the OMPL class \texttt{ob::CompoundState-\\Space} and storing the state space in the first \texttt{ob::StateSpace} element, and the hybrid time space in the second \texttt{ob::HybridTimeStateSpace} element, where the double-valued attribute \texttt{position} and integer-valued attribute \texttt{jumps}, representing the flow time and number of jumps, respectively. }
\end{revised}

\begin{revision}
\subsubsection{Inputs}
{In the \texttt{Motion} datastructure, the input is stored as an object of the OMPL class \texttt{oc::Control}. Input sets \( U_C\) and \( U_D\) have minimum and maximum values for each state, implemented as $2$-by-$m$ arrays, containing double-valued scalars, where $m$ denotes the dimension of input in~(\ref{model:generalhybridsystem}).}
\end{revision}

\subsubsection{Edge}
{In the \texttt{Motion} datastructure, the edge $e$ is implemented as a C++ pointer to the left endpoint of the edge, or the \texttt{parent}. The right endpoint of the edge is represented by the vertex $x_v$, or its attribute \texttt{state}.}

\section{Implementation of the HyRRT Algorithm} \label{hyRRT-steps}

Next, we introduce the main steps executed by HyRRT.
Given the motion planning problem \( P =
(X_0, X_f, X_u, (C, f, D, g)) \) and the input library \( (U_C, U_D) \), HyRRT performs the following steps, as in \cite{wang2022rapidly}:
\begin{steps}
\item[Step 1:] Sample a finite number of points from \( X_0 \) and
initialize a search tree \( \mathcal{T} = (V, E) \) by adding
vertices associated with each sampling point.

\item[Step 2:] Randomly select a point \( x_{\texttt{rand}} \) from \( C \)  or \( D\) by randomly sampling the state space and set checking using flow and jump sets $C$ and $D$, respectively. The definite planning space is defined in Section \ref{state-space}.

\item[Step 3:] Find the vertex \( v_{\texttt{cur}} \) associated with the state value that has minimal distance to \( x_{\texttt{rand}} \).

\item[Step 4:] Randomly select an input signal (value) from \( U_C \)
(\( U_D \)) if the flow (jump, respectively) regime is
selected. Then, compute a solution pair using the flow map $f$ or jump map $g$, starting from \( x_{v_{\texttt{cur}}} \) with the selected input applied, denoted \( \tilde{e}_{\texttt{new}} = (\phi_{\texttt{new}}, u_{\texttt{new}}) \). If, during a simulation starting from the flow regime, \( \phi_{\texttt{new}}\) intersects with the jump regime, compute an additional solution pair using the jump map from the collision vertex. Denote the final state of \( \phi_{\texttt{new}} \) as \( x_{\texttt{new}} \). If \( \tilde{e}_{\texttt{new}} \) does not intersect with \( X_u \), add a vertex \( v_{\texttt{new}} \) associated with \( x_{\texttt{new}} \) to \( V \) and an edge \( (v_{\texttt{cur}}, v_{\texttt{new}}) \) associated with \( \tilde{e}_{\texttt{new}} \) to \( E \). Then, go to Step 2. 
\end{steps}
 \ifbool{shorten}{ Following the above overview of HyRRT, the pseudocode for the proposed algorithm can be found in \cite[Algorithm 2]{wang2024motion}. }
The inputs of HyRRT are the problem \( P = (X_0, X_f, X_u, (C, f, D, g)) \), the input library \( (U_C, U_D) \), an upper bound \( K \in \mathbb{N}_{>0} \) for the number of iterations to execute, and two tunable sets \( X_c \subset C \) and \( X_d \subset D \), which act as constraints in finding the closest vertex to \( x_{\texttt{rand}} \). 
 Revisiting Example \ref{example:pinball}, we next introduce the implementation of selected steps in HyRRT.
\begin{algorithm}\footnotesize
\caption{HyRRT algorithm}
\label{algorithm:HyRRT}
\textbf{Input:} $X_0, X_f, X_u, \mathcal{H} = (C, f, D, g), (U_C, U_D), p \in (0, 1), K \in \mathbb{N} > 0$

\begin{algorithmic}[1]
\STATE $\mathcal{T} \gets \texttt{init}(X_0)$
\FOR{$k = 1$ to $K$}
    \STATE \text{randomly select a real number } $r$ \texttt{ from } $[0, 1]$
    \IF{$r \leq p$}
        \STATE $x_{\texttt{rand}} \gets \texttt{random\_state}(C)$
        \STATE \texttt{extend}($\mathcal{T}, x_{\texttt{rand}}, (U_C, U_D), \mathcal{H}, X_u, \texttt{flow}$)
    \ELSE
        \STATE $x_{\texttt{rand}} \gets \texttt{random\_state}(D)$
        \STATE \texttt{extend}($\mathcal{T}, x_{\texttt{rand}}, (U_C, U_D), \mathcal{H}, X_u, \texttt{jump}$)
    \ENDIF
\ENDFOR \RETURN$\mathcal{T}$
\end{algorithmic}
\begin{algorithmic}[1]
\Function{\texttt{extend}}{$\mathcal{T}, x, (U_C, U_D), \mathcal{H}, X_u, \texttt{flag}$} 
    \STATE $v_{\texttt{cur}} \gets \texttt{nearest\_neighbor}(x, \mathcal{T}, \mathcal{H}, \texttt{flag})$
    \IF{\texttt{new\_ state}($x, v_{\texttt{cur}}, (U_C, U_D), \mathcal{H}, X_u, x_{\texttt{new}}, \tilde{e}_{\texttt{new}}$)}
        \STATE $v_{\texttt{new}} \gets \mathcal{T}.\texttt{add\_vertex}(x_{\texttt{new}})$
        \STATE$\mathcal{T}.\texttt{add\_edge}(v_{\texttt{cur}}, v_{\texttt{new}}, \tilde{e}_{\texttt{new}})$
        \IF{$x_{\texttt{new}} == x$}
            \STATE \texttt{Reached}
        \ELSE
            \STATE \texttt{Advanced}
        \ENDIF
    \ENDIF
    \STATE \texttt{Trapped}
\end{algorithmic}
\end{algorithm}
    \subsection{$\mathcal{T}.\texttt{init}(X_0)$}
    The function call \(\mathcal{T}: \texttt{init} \) is used to initialize a search tree \(\mathcal{T}= (V, E) \). It randomly selects a finite number of points from \( X_0 \). For each sampling point \( x_0 \), a vertex \( v_0 \) associated with \( x_0 \) is added to \( V \). At this step, no edge is added to \( E \). The static function \texttt{void initTree(void)} implements this step, as shown below.
    \begin{lstlisting}[language=C++]
void oc::HyRRT::initTree(void)
{
    // get initial states with PlannerInputStates helper, pis_
    while(const ob::State *st = pis_.nextStart())
    {
        auto *motion = new Motion(si_);
        si_->copyState(motion->state, st);
        motion->root = motion->state;
        // Add start motion to the tree object nn_
        nn_->add(motion);
    }
}
\end{lstlisting}
\vspace{-0.05cm}
    \subsection{\( x_\text{rand} \leftarrow \text{random\_state}(S) \)}
  The function call \( \text{random\_state} \) randomly selects a point from the set \( S \subseteq \mathbb{R}^n \). It is designed to select from 
  \( C \) or \( D \) separately, rather than selecting from \( C \cup D \). The reason is that if \( C \) (or \( D \)) has zero measure while \( D \) (or \( C \), respectively) does not, the probability that the point selected from \( C \cup D \) lies in \( C \) (or \( D \), respectively) is zero, which would prevent establishing probabilistic completeness. The flow and jump sets $C$ and $D$ are defined as functions \texttt{flowSet\_}\footnote{Following OMPL style for class member variables, we include an underscore after \texttt{flowSet}. We treat \texttt{flowSet\_} as a member variable rather than a function because it is specific to instances of the class.} and \texttt{jumpSet\_}, respectively, and the random selection is implemented by the static function \texttt{randomSample}.
    \begin{enumerate}
    \item[1)] Jump set $D$ is implemented as the lambda function \texttt{jumpSet\_}. It takes in an arbitrary state as an input and outputs \texttt{true} if the state belongs to jump set $D$, and \texttt{false} if not. 

Below, we demonstrate how to implement the jump set for the pinball example.
\begin{lstlisting}[language=C++]
bool jumpSetExample(oc::HyRRT::Motion *motion)
{
    // A structure containing the coordinates of the different pinball paddles
    PinballSetup pinballSetup;

    // Extract the x-components of the state's position and velocity
    auto *motion_state = motion->state->as<ob::CompoundState>()->as<ob::RealVectorStateSpace::StateType>(0);
    double x1 = motion_state->values[0];
    double v1 = motion_state->values[2];

    for (std::vector<double> paddleCoord : pinballSetup.paddleCoords)    // If ball is in any of the paddles
    {   // inPaddle is a helper function that has been defined in the pinball example file
        if (inPaddle(motion, paddleCoord))
            return true;
    }

    if ((x1 <= 0 && v1 < 0) || (x1 >= 5 && v1 > 0)) // If ball is in either side wall
        return true;

    return false;
}
\end{lstlisting}
    \item[2)]  Flow set $C$ is implemented as the lambda function \texttt{flowSet\_}. It takes in an arbitrary state as an input and outputs \texttt{true} if the state belongs to flow set $C$, and \texttt{false} if not. 
    Below, we demonstrate how to implement the flow set for the pinball example.
\begin{lstlisting}[language=C++]
bool flowSet(oc::HyRRT::Motion *motion)
{
    return !jumpSet(motion);
}
\end{lstlisting}
\item[3)] The OMPL \texttt{SpaceInformation} class's built-in state space sampler function \texttt{randomSample} is utilized to select a random state.
\begin{lstlisting}[language=C++]
void oc::HyRRT::randomSample(Motion *randomMotion)
{
    ob::StateSamplerPtr sampler_ = si_->allocStateSampler();
    sampler_->sampleUniform(randomMotion->state);
}
\end{lstlisting}
\vspace{-0.05cm}
\end{enumerate}
    \subsection{ \( v_\text{cur} \leftarrow \text{nearest\_neighbor}(x_\text{rand}, \mathcal{T}, H, \text{flag}) \)}
   HyRRT searches for a vertex \( v_\texttt{cur} \) in the search tree \(\mathcal{T}= (V, E) \) such that its associated state value has minimal distance to \( x_\texttt{rand} \). This process is implemented as follows.
    \begin{itemize}
        \item When $x_\texttt{rand} \in C$, the following optimization problem is solved over \( X_c \):
        \begin{problem}
         $
             \quad \min_{v \in V} \| x_v - x_\texttt{rand} \| \quad \texttt{s.t.} \quad x_v \in X_c.
         $ 
         \end{problem}
        \item When $x_{rand} \in D$, the following optimization problem is solved over \( X_d \):
        \begin{problem}
         $
         \quad \min_{v \in V} \| x_v - x_\texttt{rand} \| \quad \texttt{s.t.} \quad x_v \in X_d.
         $
         \end{problem}
    \end{itemize}
   The data of Problem 2 and Problem 3 comes from the arguments of the \( \texttt{nearest\_neighbor} \) function call. This optimization problem can be solved by traversing all the vertices in \(\mathcal{T}= (V, E) \). The unsafe set is defined as the function \texttt{unsafeSet\_}. Below, we present the static optimization function used to solve Problems 2 and 3. 
   \begin{enumerate}
           \item[1)] \texttt{setDistanceFunction}: Set the function \texttt{distanceFunc\_} that computes distance between states. 
Function is default-initialized to calculate Euclidean distance.
\begin{revised}
\begin{lstlisting}[language=C++]
double distanceFunction(ob::State *phi1, ob::State *phi2) {...}
cHyRRT.setDistanceFunction(distanceFunction);
\end{lstlisting}
\end{revised}
        \item[2)] \texttt{NearestNeighbors:} A built-in OMPL class is used to store and search within the search tree of \texttt{Motion} objects \texttt{nn\_}, which is an object of \\ \texttt{NearestNeighbors}. The function call \texttt{nearest} solves both Problem 2 and 3, returning the \texttt{Motion} in \texttt{nn\_} with the minimal distance to \texttt{randomMotion}, measured by \texttt{distanceFunc\_}.
\begin{lstlisting}[language=C++]
std::shared_ptr<NearestNeighbors<Motion *>> nn_;
nn_->nearest(randomMotion);
\end{lstlisting}

    \item[3)] \texttt{$X_u$:} The unsafe set $X_u$ is implemented as the lambda function \texttt{unsafeSet\_}. It takes in an arbitrary state as an input and outputs \texttt{true} if the state belongs to unsafe set $X_u$ and \texttt{false} if not. 
    Below, we demonstrate how to implement the unsafe set for the pinball example.
\begin{lstlisting}[language=C++]
bool unsafeSet(oc::HyRRT::Motion *motion)
{
    auto *motion_state = motion->state->as<ob::CompoundState>()->as<ob::RealVectorStateSpace::StateType>(0);
    double x1 = motion_state->values[0];
    double x2 = motion_state->values[1];

    if (((x1 >= 0 && x1 < 2) || (x1 > 3 && x1 <= 5)) && x2 <= -10)
        return true;
    return false;
}
\end{lstlisting}
\end{enumerate}
    \subsection{ \( \text{return} \leftarrow \text{new\_state}(x_\text{rand}, v_\text{cur}, (U_C, U_D), H, X_{unsafe}, x_\text{new}, \\ \tilde{e}_\text{new}) \)}
    If \( x_{v_\texttt{cur}} \in C \setminus D \) (or \( x_{v_\texttt{cur}} \in D \setminus C \)), a new solution pair \( \tilde{e}_\texttt{new} \) to the hybrid system \( H \), starting from \( x_{v_\texttt{cur}} \), is generated by applying an input signal \( \tilde{u} \) (or an input value \( u_D \)) randomly selected from \( U_C \) (or \( U_D \), respectively). If \( x_{v_\texttt{cur}} \in C \cap D \), then this function generates \( \tilde{e}_\texttt{new} \) by randomly selecting flows or jumps. The final state of \( \tilde{e}_\texttt{new} \) is denoted as \( x_\texttt{new} \).
   Note that the choices of inputs are random. After \( \tilde{e}_\texttt{new} \) and \( x_\texttt{new} \) are generated, HyRRT checks if there exists \( (t, j) \in \texttt{dom}(\tilde{e}_\texttt{new}) \) such that \( \tilde{e}_\texttt{new}(t, j) \in X_u \). If so, then \( \tilde{e}_\texttt{new} \) intersects with the unsafe set, no $x_{new}$ is returned. Otherwise, $x_{new}$ is successfully returned. When $x_{v_{cur}} \in C$, this step is implemented by the function \texttt{continuousSimulator\_}, which uses numerical integration to propagate the state, given the randomly generated flow time input, and with respect to the dynamics of flow map $f$. Then, the solution pair is checked for collision with the jump regime using the function \texttt{collisionChecker\_}. When $x_{v_{cur}} \in D$ or $x_{v_{cur}}$ experiences a collision with the jump regime during continuous propagation, a jump is simulated by the discrete simulator \texttt{discreteSimulator\_}.   
   \begin{enumerate}
    \item[1)] \nw{Flow map $f$ an object of the OMPL class \texttt{oc::StatePropagator}\footnote{See OMPL class reference: https://ompl.kavrakilab.org/classompl\_1\_1\newline control\_1\_1StatePropagator.html}.} Each discretized integration step spans a duration specified by the user as class member variable \texttt{flowStepDuration\_}. A full example implementation of flow map $f$ can be found in \cite{xu2024chyrrtchysstmotionplanning}.   
    \item Next, we present the functions used to define the maximum propagation duration and discretized integration step duration: 
The applied control input $u$ is randomly selected from control input set $U_C$. 
    \begin{enumerate}
     \item[a)] \texttt{setTm}: Set the maximum flow time \texttt{Tm\_}, which must be positive.
    \item[b)] \texttt{setFlowStepDuration}: Set the flow time for a given integration step \texttt{flowStepDuration\_}, which must be positive and less than or equal to the maximum flow time \texttt{Tm\_}.
    \item[c)] \texttt{setFlowInputRange}: Set the vectors of minimum and maximum input values for integration in the flow regime. Minimum input values must be less than or equal to their corresponding maximum input values.
\end{enumerate}
Below, we demonstrate the declaration of the \texttt{oc::StatePropagator} object used by the internal function \texttt{continuousSimulator\_} for the pinball example.  
\begin{lstlisting}[language=C++]
void flowODE(const oc::ODESolver::StateType &q, const oc::Control *c,
             oc::ODESolver::StateType &qdot)
{ /* Modify qdot... */ }

// Initialize oc::StatePropagator, an attribute of si, for use in continuousSimulator_
... 
// Define state space as hybridSpacePtr using procedure from Section III-B.1
... 

// Define flow control space; duplicate for jump control space
oc::RealVectorControlSpace *flowControlSpace = new oc::RealVectorControlSpace(hybridSpacePtr, 2);

// Define control space
oc::RealVectorControlSpace *flowControlSpace = new oc::RealVectorControlSpace(hybridSpacePtr, 2);
oc::RealVectorControlSpace *jumpControlSpace = new oc::RealVectorControlSpace(hybridSpacePtr, 2);

// Repeat for jumpControlSpace
ompl::base::RealVectorBounds flowBounds(1);
flowBounds.setLow(0, -0.5);
flowBounds.setHigh(1, 1);
flowBounds.setLow(0, -1);
flowBounds.setHigh(1, 1);
flowControlSpace->setBounds(flowBounds);

oc::RealVectorControlUniformSampler flowControlSampler(flowControlSpace);
flowControlSpace->setControlSamplerAllocator([flowControlSpace](const oc::ControlSpace *space) -> oc::ControlSamplerPtr {
    return std::make_shared<oc::RealVectorControlUniformSampler>(space);
});

oc::ControlSpacePtr flowControlSpacePtr(flowControlSpace);

oc::CompoundControlSpace *controlSpace = new oc::CompoundControlSpace(hybridSpacePtr);
controlSpace->addSubspace(flowControlSpacePtr);
controlSpace->addSubspace(jumpControlSpacePtr);
oc::ControlSpacePtr controlSpacePtr(controlSpace);

// Construct a space information instance for this state space
oc::SpaceInformationPtr si(new oc::SpaceInformation(hybridSpacePtr, controlSpacePtr));
oc::ODESolverPtr odeSolver (new oc::ODEBasicSolver<> (si, &flowODE));
...

// Allocate flow input control sampler; replicate for jump control sampler
oc::RealVectorControlUniformSampler flowControlSampler(flowControlSpace);
flowControlSpace->setControlSamplerAllocator([flowControlSpace](const oc::ControlSpace *space) -> oc::ControlSamplerPtr {
    return std::make_shared<oc::RealVectorControlUniformSampler>(space);
});

oc::ControlSpacePtr flowControlSpacePtr(flowControlSpace);
oc::ControlSpacePtr jumpControlSpacePtr(jumpControlSpace);

oc::CompoundControlSpace *controlSpace = new oc::CompoundControlSpace(hybridSpacePtr);
controlSpace->addSubspace(flowControlSpacePtr);
controlSpace->addSubspace(jumpControlSpacePtr);

oc::ControlSpacePtr controlSpacePtr(controlSpace);

// Construct a space information instance for this state space
oc::SpaceInformationPtr si(new oc::SpaceInformation(hybridSpacePtr, controlSpacePtr));

si->setStatePropagator(oc::ODESolver::getStatePropagator(odeSolver));
si->setPropagationStepSize(0.01);
si->setup();
oc::HyRRT cHyRRT(si);
\end{lstlisting}

The same approach to instantiating the tool on lines~55-60, where an instance of \texttt{RealVectorStateSpace}, defining the state space, is passed into an instance of \texttt{SpaceInformation}, and subsequently into \texttt{HyRRT}, can be applied to both motion planning algorithms presented in this paper. The parameter definition method on line 58 can be generalized for the remaining customizable parameters by replacing \texttt{Tm} with the desired parameter name, except for the control input ranges. 
\revision{The method used to define $U_C$ on lines~40-43 can be generalized to $U_D$, by replacing the arguments of \texttt{setLow} and \texttt{setHigh} with the corresponding limits of $U_D$.}
    \item[2)] The discrete simulator is implemented within the lambda function \\ \texttt{discreteSimulator\_}. The function \texttt{discreteSimulator\_} returns a state propagated once. 
    \item The control input is defined as $u \in U_D$. 
Below, we demonstrate the implementation of \texttt{discreteSimulator\_} for the pinball example.    
\begin{lstlisting}[language=C++]
ob::State *discreteSimulator(ob::State *x_cur, std::vector<double> u, ob::State *new_state)
{
    // Modify new_state...
}
\end{lstlisting}
    \item[3)] The collision checker is implemented as the lambda function \\ \texttt{collisionChecker\_}. It takes in an input of a \texttt{Motion} object and, if a collision occurs, outputs \texttt{true} and updates the edge's right endpoint to reflect the collision state. If no collision occurs, then the function outputs \texttt{false} and makes no modification to the edge.  
By default, the function is a point-by-point collision checker that checks each point using the \texttt{jumpSet\_}. Below, we demonstrate the implementation of the collision checker for a general example.
\begin{lstlisting}[language=C++]
bool collisionChecker(std::vector<ob::State *> *solutionPair, std::function<bool(ob::State *state)> obstacleSet, ob::State *new_state) {
    // Modify state if needed
    ...
    return false;
}
\end{lstlisting}
\end{enumerate}
    \subsection{ \( v_\texttt{new} \leftarrow\mathcal{T}.\texttt{add\_vertex}(x_\texttt{new}) \) and \(\mathcal{T}.\texttt{add\_edge}(v_\texttt{cur}, v_\texttt{new}, \tilde{e}_\texttt{new}) \)}
    The function call \(\mathcal{T}.\texttt{add\_vertex}(x_\texttt{new}) \) adds a new vertex \( v_\texttt{new} \) associated with \( x_\texttt{new} \) to \(\mathcal{T}\) and returns \( v_\texttt{new} \). The function call \(\mathcal{T}.\texttt{add\_edge}(v_\texttt{cur}, v_\texttt{new}, \tilde{e}_\texttt{new}) \) adds a new edge \( e_\texttt{new} = (v_\texttt{cur}, v_\texttt{new}) \) associated with \( \tilde{e}_\texttt{new} \) to \(\mathcal{T}\).
    
\subsection{Solution Checking Process:}
A solution checking function is employed to check if a path in \(\mathcal{T}\) can be used to construct a motion plan for the given motion planning problem. If this function finds a path \( p = ((v_0, v_1), (v_1, v_2), \dots, (v_{n-1}, v_n)) =: (e_0, e_1, \dots, e_{n-1}) \) in \(\mathcal{T}\) such that 1) \( x_{v_0} \in X_0 \) and 2) \( x_{v_n} \in X_f \), then the solution pair $\tilde{p}$ is a motion plan for the given motion planning problem. This function is constructed by implementing the abstract OMPL class ob::Goal\footnote{See OMPL class reference: https://ompl.kavrakilab.org/classompl\_1\_1\\base\_1\_1Goal.html}.

\section{Implementation of the HySST Algorithm}
HySST generates asymptotically near-optimal solutions, with only two notable deviations from the main steps of HyRRT in Section \ref{hyRRT-steps}:

\begin{algorithm}[htbp]
\caption{HySST algorithm}
\label{algorithm:HySST}
\textbf{Input:} $X_0, X_f, X_u, \mathcal{H} = (C, f, D, g), (U_C, U_D), p_n \in (0, 1), K \in \mathbb{N}_{>0}, X_c, X_d, \newline \epsilon_{BN}, \epsilon_s$
\begin{algorithmic}[1]
    \STATE $\mathcal{T} \gets \texttt{init}(X_0)$
    \STATE $V_{\texttt{active}} \gets V$, $V_{\texttt{inactive}} \gets \emptyset$, $S \gets \emptyset$
    \FOR{all $v_0 \in V$}
        \IF{is\_vertex\_locally\_the best($x_{v_0}$, 0, $S$, $\epsilon_s$)}
            \STATE \parbox[t]{.9\linewidth}{
                $(S, V_{\texttt{active}}, V_{\texttt{inactive}}, E) \gets$ \texttt{prune\_ dominated\_vertices}($v_0$, $S$, $V_{\texttt{active}}$, $V_{\texttt{inactive}}$, $E$)}
        \ENDIF
    \ENDFOR
    \FOR{$k = 1$ \textbf{to} $K$}
        \STATE randomly select a real number $r$ from $[0, 1]$
        \IF{$r \leq p$}
            \STATE $x_{\texttt{rand}} \gets \texttt{random\_state}(C)$
            \STATE $v_{\texttt{cur}} \gets \texttt{best\_near\_selection}(x_{\texttt{rand}}, V_{\texttt{active}}, \newline \epsilon_{BN}, X_c)$
        \ELSE
            \STATE $x_{\texttt{rand}} \gets \texttt{random\_state}(D)$
            \STATE $v_{\texttt{cur}} \gets \texttt{best\_near\_selection}(x_{\texttt{rand}}, V_{\texttt{active}}$, $\epsilon_{BN}$, $X_d)$
        \ENDIF
        \STATE \parbox[t]{.9\linewidth}{
            $(\texttt{is\_a\_new\_vertex\_generated}$, $x_{\texttt{new}}$, $\tilde{e}_{\texttt{new}}$, $\texttt{cost}_{\texttt{new}}) \gets$$\texttt{new\_state}(v_{\texttt{cur}}, (U_C, U_D), \mathcal{H}, X_u)$}
        \IF{$\texttt{is\_a\_new\_vertex\_generated} \textbf{ and}$ is\_vertex\_locally\_the\_best($x_{\texttt{new}}$, $\texttt{cost}_{\texttt{new}}$, S, $\epsilon_s$)}
            \STATE \parbox[t]{.9\linewidth}{
                $v_{\texttt{new}} \gets V_{\texttt{active}}.\texttt{add\_vertex}(x_{\texttt{new}}, \texttt{cost}_{\texttt{new}})$ \\
                \hspace*{1.5em} $E.\texttt{add\_edge}(v_{\texttt{cur}}, v_{\texttt{new}}, \tilde{e}_{\texttt{new}})$}
            \STATE \parbox[t]{.9\linewidth}{
                $(S, V_{\texttt{active}}, V_{\texttt{inactive}}, E) \gets$ \\
                \hspace*{1.5em} $\texttt{prune\_dominated\_vertices}$$(v_{\texttt{new}}$, $S$, $V_{\texttt{active}}, V_{\texttt{inactive}}, E)$}
        \ENDIF
    \ENDFOR
    \STATE \textbf{return}$ \; \mathcal{T}$
\end{algorithmic}
\end{algorithm}
\begin{steps}
    \item[Step 3:] Find the vertex \( v_{\texttt{cur}} \) associated with the state value that has the minimal cost functional, within the neighborhood defined by a ball of selection radius $\epsilon_{BN}$ of \( x_{\texttt{rand}} \). If no vertex exists within the neighborhood, the nearest vertex in $V$ is selected. \nw{The larger the magnitude of selection radius $\epsilon_{BN}$, the stronger the preference is for lower cost vertices, and thus, the lower the average cost for generated solutions. Typically, however, the greater the preference for lower cost vertices, the greater the computational burden.} The selection radius $\epsilon_{BN}$ is set using \texttt{setSelectionRadius(double selectionRadius)} function.
    \begin{revision}
        \begin{enumerate}
            \item[1)] \texttt{setSelectionRadius(double selectionRadius)}: Set the scalar value $\epsilon_{BN}$ used to select vertex closest to the randomly sampled vertex. Must be greater than or equal to zero.
                \begin{lstlisting}[language=C++]
double selectionRadius = ...;
cHySST.setSelectionRadius(selectionRadius); \end{lstlisting}
        \end{enumerate}
    \item[Step 4:] Once a solution pair is computed, if  \( \tilde{e}_{\texttt{new}} \) does not intersect with \( X_u \) and  \( v_{\texttt{new}} \) has a minimal cost within the neighborhood defined by a ball of pruning radius $\epsilon_{S}$,  add the vertex \( v_{\texttt{new}} \) associated with \( x_{\texttt{new}} \) to \( V \) and an edge \( (v_{\texttt{cur}}, v_{\texttt{new}}) \) associated with \( \tilde{e}_{\texttt{new}} \) to \( E \). Then, the pruning process removes from the search tree the vertices without child vertices and with higher cost, in the neighborhood of the new vertex \( v_{\texttt{active}} \) defined by a ball of radius $\epsilon_{S}$. This additional step maintains \( S \), a static set of witnesses to sparsify the vertices. \nw{The larger the magnitude of pruning radius $\epsilon_{S}$, the sparser the tree, and the worse the cost, compared to that of the real optimal solution.} 
    Below, we define the pruning radius $\epsilon_{S}$ through \texttt{setPruningRadius(double pruningRadius)}.
    , which sets the scalar value  $\epsilon_{S}$ used to define a surrounding region for removing representative vertices from the witness set. 
    The value must be zero or positive.
                \begin{lstlisting}[language=C++]
double pruningRadius = ...;
cHySST.setPruningRadius(pruningRadius); \end{lstlisting}
        \end{revision}
\end{steps}
The HySST algorithm pseudocode can be found in \cite{wang2023hysst}.

The number of solutions allowed in a single instance of HySST is set using the  \texttt{setBatchSize(int batchSize)} function. Once the maximum number of solutions is reached, the tool's \texttt{solve} function, which generates the solution path, will return the solution with the lowest cost. Batch size is default-initialized to be $1$ and must be a positive integer if customized. The principle and effect of batch size will be further discussed in Section \ref{section:discussion} through Example \ref{example:pinball}.
\begin{lstlisting}[language=C++]
double batchSize = ...;
cHyRRT.setBatchSize(batchSize);
\end{lstlisting}

\section{{Simulation Results}}
We illustrate both cHyRRT and cHySST in Example \ref{example:pinball}. 
\subsection{Modified Pinball Game (revisited)}
    The following procedure is used to generate a solution to the motion planning problem for the specific instance of the pinball system presented in Example \ref{example:pinball} using cHySST: 
    \begin{enumerate}
        \item Inside the C++ file \texttt{SSTPinballPlanning.cpp}, the cost function, initial conditions, and planning and simulation parameters are set; the cost of a state $x$ of the pinball is the negative value of its total distance traveled along the $x$-axis, from its starting state $x_0$ to its current state $x$. 
        \item The motion planner is run using the \texttt{solve} function within cHySST to return a trajectory. 
    \end{enumerate}
    \nw{On average, the pinball example experienced a reduction in computation time, from an average of $6.425$ seconds per run when using HyRRT in MATLAB, to an average of $0.7173$ seconds per run when using cHyRRT (C++). }

    \subsection{Collision-resilient Tensegrity Multicopter}
{A simulated solution to a collision-resilient tensegrity multicopter in the horizontal plane that can operate after colliding with a concrete wall is shown, where the position of the multicopter along the $y$-axle of the ball is plotted as a function of the position along the $x$-axle.}

 \begin{figure}[htbp]
        \subfigure[Simulation result solved by cHySST with initial state $x_0=(1, 2, -1, 0)$.]{\includegraphics[width=0.35\columnwidth]{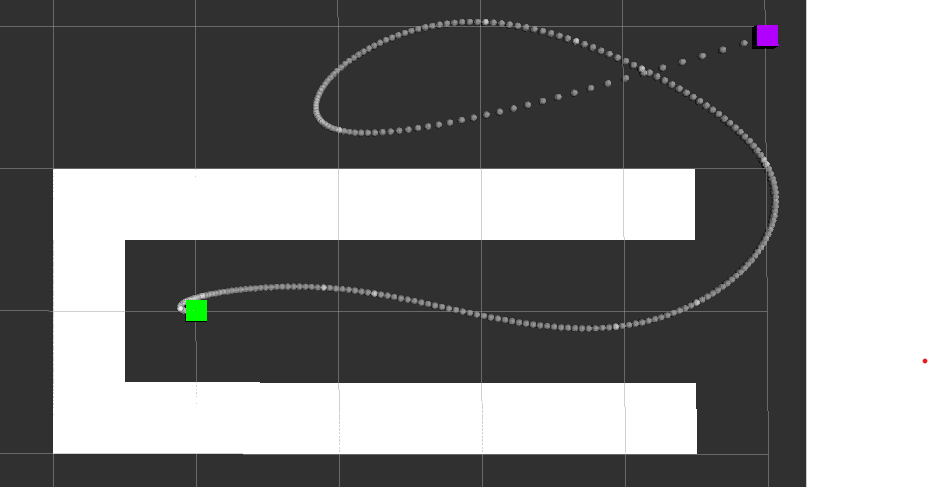}}       
        \subfigure[Simulation result solved by cHyRRT with initial state $x_0=(3, 2, 0, 0)$.]{\includegraphics[width=0.33\columnwidth]{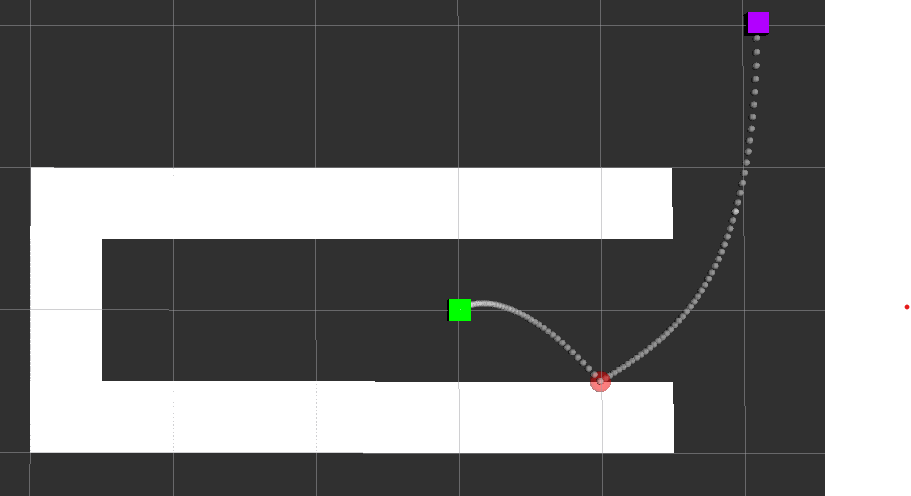}}
        \subfigure[Simulation result solved by cHyRRT with initial state $x_0=(0.55, 2, 0, 0)$.]{\includegraphics[width=0.3\columnwidth]{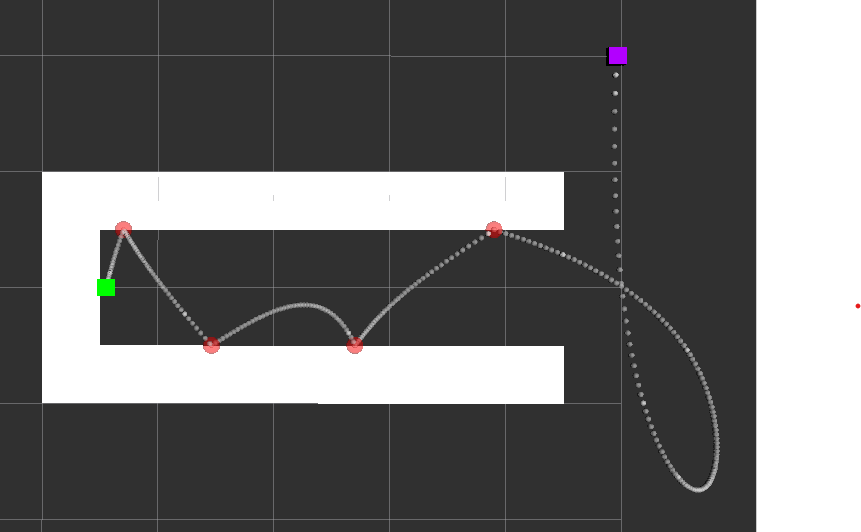}} 
        \vspace{-0.3cm}
        \caption{\revised{Simulated solution to the multicopter example, graphed as $x_1$ vs $x_2$. The start and goal vertices are marked by green and purple squares, respectively, and vertices in the jump regime are marked by red circles.}}
        \label{fig:multicopter}
    \end{figure}
    
{As previously defined in \cite{wang2023hysst}, the state of the multicopter is composed of the position vector $p := (p_x, p_y) \in \mathbb{R}^2$, where $p_x$ denotes the position along the $x$-axis and $p_y$ denotes the position along the $y$-axis, the velocity vector $v := (v_x, v_y) \in \mathbb{R}^2$, where $v_x$ denotes the velocity along the $x$-axis and $v_y$ denotes the velocity along the $y$-axis, and the acceleration vector $a := (a_x, a_y) \in \mathbb{R}^2$ where $a_x$ denotes the acceleration along the $x$-axis and $a_y$ denotes the acceleration along the $y$-axis. The state of the system is $x := (p, v, a) \in \mathbb{R}^6$ and its input is $u := (u_x, u_y) \in \mathbb{R}^2$. The environment is assumed to be known. Define the region of the walls as $W \subset \mathbb{R}^2$, represented by white rectangles in Figure \ref{fig:multicopter}.}

{Flow is allowed when the multicopter is in the free space. Hence, the flow set is $C := \{((p, v, a), u) \in \mathbb{R}^6 \times \mathbb{R}^2 : p \notin W\}$. The dynamics of the quadrotors when no collision occurs can be captured using time-parameterized polynomial trajectories because of its differential flatness as \cite{8206119} 
\[
\dot{x} =
\begin{bmatrix}
v \\
a \\
u
\end{bmatrix}
=: f(x, u) \quad (x, u) \in C.
\]
Note that the post-collision position stays the same as the pre-collision position. Therefore, $p^+ = p$. Denote the velocity component of $v = (v_x, v_y)$ that is normal to the wall as $v_n$ and the velocity component that is tangential to the wall as $v_t$. Then, the velocity component $v_n$ after the jump is modeled as $v_n^+ = -e v_n =: g_{v_n}(v)$ where $e \in (0, 1)$ is the coefficient of restitution. The velocity component $v_t$ after the jump is modeled as $v_t^+ = v_t + \kappa(-e - 1) \arctan \left(\frac{v_t}{v_n}\right) =: g_{v_t}(v)$ where $\kappa \in \mathbb{R}$ is a constant; see [14]. Denote the projection of the updated vector $(v_n^+, v_t^+)$ onto the $x$-axis as $\overline{x}(v_n^+, v_t^+)$ and the projection of the updated vector $(v_n^+, v_t^+)$ onto the $y$-axis as $\overline{y}(v_n^+, v_t^+)$. Therefore,
\[
v^+ =
\begin{bmatrix}
\overline{x}(g_{v_n}(v), g_{v_t}(v)) \\
\overline{y}(g_{v_n}(v), g_{v_t}(v))
\end{bmatrix}
=: g_v(v).
\]
We assume that $a^+ = 0$. The discrete dynamics capturing the collision process is modeled as
\[
x^+ =
\begin{bmatrix}
p \\
g_v(v) \\
0
\end{bmatrix}
=: g(x, u) \quad (x, u) \in D.
\]
The jump is allowed when the multicopter is on the wall surface with positive velocity towards the wall. Hence, the jump set is
\[
D := \{((p, v, a), u) \in \mathbb{R}^6 \times \mathbb{R}^2 : p \in \partial W, v_n < 0\}.
\]
Given the initial state set as $X_0 = \{(1, 2, 0, 0, 0, 0)\}$, the final state set as $X_f = \{(5, 4)\} \times \mathbb{R}^4$, and the unsafe set as
\begin{multline}
X_unsafe = \{(x, u) \in \mathbb{R}^6 \times \mathbb{R}^2 : p_x \in (-\infty, 0] \cup [6, \infty), \\
p_y \in (-\infty, 0] \cup [5, \infty), (p_x, p_y) \in \text{int}\, W\},
\end{multline}
{On average, the multicopter example experienced a reduction in computation time, from an average of $4.89$ seconds per run when using HySST in MATLAB, to an average of $0.779$ seconds per run when using cHySST (C++). Using SST, an algorithm designed for purely continuous systems,  to generate a motion plan under a computation time limit of 30 seconds for the multicopter example resulted in valid motion plans for only 5 out of 30 trials, compared to 27 out of 30 trials when using HySST. By considering motion plans exhibiting hybrid dynamical behavior, HySST is able to explore more feasible solutions, thus achieving higher success rates than SST.}
}
\vspace{-0.0cm}
\section{Discussion: Evolution of Solution Costs Over Range of HySST Solution Batch Sizes and Limitations}
\label{section:discussion}
\subsection{Evolution of Solution Costs Over Range of HySST Solution Batch Sizes}
The data collected from $15$ runs of cHySST solving the pinball motion planning problem shows an inverse correlation between solution batch size and the cost of the lowest-cost solution. As the batch size increases, both the minimum and mean costs of the $15$ lowest-cost solutions decrease.
However, while increasing batch size generally reduces the cost of the generated solution with the best cost, it also increases computational consumption. As more of the planning space is explored and vertices are sparsified, the rate at which new vertices $v_{new}$ are added to $V$ declines, increasing the difficulty of finding a unique, lower-cost solution.
\begin{figure}[H]
    \centering
    \includegraphics[width=0.6\linewidth]{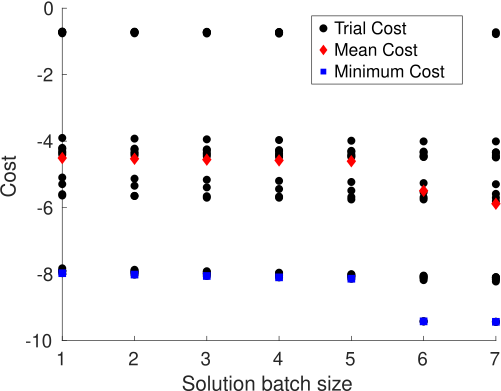}
    \vspace{-0.2cm}
    \caption{Cost of lowest-cost solution given a solution batch size.}
    \label{fig:pinball-cost-trend}
\end{figure}
\vspace{-0.4cm}

\subsection{Limitations} \begin{nw}
{Both HySST and HyRRT face two major limitations, which pose opportunities for promising future work: i) dependence on a successfully formulated hybrid model and ii) dimensional explosion. Expertise in hybrid systems is required to develop an accurate hybrid model. Both algorithms also face exponential growth of vertices during search. Let $n$ be the number of input signals in inputs sets $U_C$ and $U_D$. While a solution is yet to be found, $n$ input signals can be applied to $m$ vertices, with the number of vertices $m$ increasing as the search tree grows, resulting in dimensional explosion. A more detailed time complexity analysis for HyRRT and HySST can be found in \cite[Section 3.4]{time-complexity-analysis}.
}
\end{nw}

\section{Conclusion}
The two tools cHyRRT and cHySST, for planning of hybrid systems, were described and illustrated in examples. Leveraging the computational efficiency of C++, the applicability to high-dimensional hybrid systems of the RRT-type and SST-type tools, and the generalizability of OMPL and ROS to robotics applications, we present two motion planning tools.
In future work, we will implement HyRRT-Connect, a bidirectional RRT algorithm, in C++/OMPL.

\bibliography{reference}

\begin{thebibliography}{15}
\providecommand{\natexlab}[1]{#1}
\providecommand{\url}[1]{\texttt{#1}}
\expandafter\ifx\csname urlstyle\endcsname\relax
  \providecommand{\doi}[1]{doi: #1}\else
  \providecommand{\doi}{doi: \begingroup \urlstyle{rm}\Url}\fi

\bibitem[Branicky et~al.(2003)Branicky, Curtiss, Levine, and Morgan]{branicky2003samplingbased}
Michael~S Branicky, Michael~M Curtiss, John Levine, and Scott Morgan.
\newblock Sampling-based planning and control.
\newblock In \emph{Proceedings of the 12th Yale Workshop on Adaptive and Learning Systems}, New Haven, CT, 2003. Citeseer.

\bibitem[Goebel et~al.(2009)Goebel, Sanfelice, and Teel]{goebel2009hybrid}
Rafal Goebel, Ricardo~G Sanfelice, and Andrew~R Teel.
\newblock Hybrid dynamical systems.
\newblock \emph{IEEE Control Systems Magazine}, 29\penalty0 (2):\penalty0 28--93, 2009.

\bibitem[Karaman and Frazzoli(2011)]{hysst-8}
Sertac Karaman and Emilio Frazzoli.
\newblock Sampling-based algorithms for optimal motion planning.
\newblock \emph{The International Journal of Robotics Research}, 30\penalty0 (7):\penalty0 846--894, 2011.
\newblock \doi{10.1177/0278364911406761}.
\newblock URL \url{https://doi.org/10.1177/0278364911406761}.

\bibitem[LaValle(1998)]{LaValle1998RapidlyexploringRT}
Steven~M. LaValle.
\newblock Rapidly-exploring random trees : a new tool for path planning.
\newblock \emph{The annual research report}, 1998.
\newblock URL \url{https://api.semanticscholar.org/CorpusID:14744621}.

\bibitem[Li et~al.(2016)Li, Littlefield, and Bekris]{hysst-9}
Yanbo Li, Zakary Littlefield, and Kostas~E. Bekris.
\newblock Asymptotically optimal sampling-based kinodynamic planning.
\newblock \emph{The International Journal of Robotics Research}, 35\penalty0 (5):\penalty0 528--564, 2016.
\newblock \doi{10.1177/0278364915614386}.
\newblock URL \url{https://doi.org/10.1177/0278364915614386}.

\bibitem[Liu et~al.(2017)Liu, Atanasov, Mohta, and Kumar]{8206119}
Sikang Liu, Nikolay Atanasov, Kartik Mohta, and Vijay Kumar.
\newblock Search-based motion planning for quadrotors using linear quadratic minimum time control.
\newblock In \emph{2017 IEEE/RSJ International Conference on Intelligent Robots and Systems (IROS)}, pages 2872--2879, 2017.
\newblock \doi{10.1109/IROS.2017.8206119}.

\bibitem[Macenski et~al.(2022)Macenski, Foote, Gerkey, Lalancette, and Woodall]{doi:10.1126/scirobotics.abm6074}
Steven Macenski, Tully Foote, Brian Gerkey, Chris Lalancette, and William Woodall.
\newblock Robot operating system 2: Design, architecture, and uses in the wild.
\newblock \emph{Science Robotics}, 7\penalty0 (66):\penalty0 eabm6074, 2022.
\newblock \doi{10.1126/scirobotics.abm6074}.
\newblock URL \url{https://www.science.org/doi/abs/10.1126/scirobotics.abm6074}.

\bibitem[Nechushtan et~al.(2010)Nechushtan, Raveh, and Halperin]{hysst-7}
Oren Nechushtan, Barak Raveh, and Dan Halperin.
\newblock Sampling-diagram automata: A tool for analyzing path quality in tree planners.
\newblock volume~68, pages 285--301, 01 2010.
\newblock ISBN 978-3-642-17451-3.
\newblock \doi{10.1007/978-3-642-17452-0_17}.

\bibitem[{\c{S}}ucan et~al.(2012){\c{S}}ucan, Moll, and Kavraki]{sucan2012the-open-motion-planning-library}
Ioan~A. {\c{S}}ucan, Mark Moll, and Lydia~E. Kavraki.
\newblock The {O}pen {M}otion {P}lanning {L}ibrary.
\newblock \emph{{IEEE} Robotics \& Automation Magazine}, 19\penalty0 (4):\penalty0 72--82, 12 2012.
\newblock \doi{10.1109/MRA.2012.2205651}.
\newblock \url{https://ompl.kavrakilab.org}.

\bibitem[Wang(2025)]{time-complexity-analysis}
Nan Wang.
\newblock Provably-correct and efficient motion planning for hybrid dynamical systems, Jan 2025.
\newblock URL \url{https://escholarship.org/uc/item/07s8n964}.

\bibitem[Wang and Sanfelice(2022)]{wang2022rapidly}
Nan Wang and Ricardo~G Sanfelice.
\newblock A rapidly-exploring random trees motion planning algorithm for hybrid dynamical systems.
\newblock In \emph{2022 IEEE 61st Conference on Decision and Control (CDC)}, pages 2626--2631. IEEE, 2022.

\bibitem[Wang and Sanfelice(2023)]{wang2023hysst}
Nan Wang and Ricardo~G Sanfelice.
\newblock {HySST}: An asymptotically near-optimal motion planning algorithm for hybrid systems.
\newblock In \emph{2023 62nd IEEE Conference on Decision and Control (CDC)}, pages 2865--2870. IEEE, 2023.

\bibitem[Wang and Sanfelice(2024)]{wang2024motion}
Nan Wang and Ricardo~G Sanfelice.
\newblock Motion planning for hybrid dynamical systems: Framework, algorithm template, and a sampling-based approach.
\newblock \emph{arXiv preprint arXiv:2406.01802}, 2024.

\bibitem[Xu et~al.(2024)Xu, Wang, and Sanfelice]{xu2024chyrrtchysstmotionplanning}
Beverly Xu, Nan Wang, and Ricardo Sanfelice.
\newblock chyrrt and chysst: Two motion planning tools for hybrid dynamical systems, 2024.
\newblock URL \url{https://arxiv.org/abs/2411.11812}.

\bibitem[Yang(2019)]{hysst-6}
Yajue Yang.
\newblock Survey of optimal motion planning.
\newblock \emph{IET Cyber-Systems and Robotics}, 1:\penalty0 13--19(6), 6 2019.
\newblock URL \url{https://digital-library.theiet.org/content/journals/10.1049/iet-csr.2018.0003}.

\end{thebibliography}
\end{document}